%% file: main.tex
\documentclass[runningheads]{llncs}
\usepackage{graphicx}
\usepackage{amsmath,amssymb} 
\usepackage{style}

\begin{document}
\title{Camera Calibration without Camera Access - \newline A Robust Validation Technique for Extended PnP Methods}


\titlerunning{A Robust Validation Technique for Extended PnP Methods}
\author{Emil Brissman\inst{1,2}\orcidID{0000-0002-0418-9694} \and
Per-Erik Forss\'en\inst{1}\orcidID{0000-0002-5698-5983} \and
Johan Edstedt\inst{1}\orcidID{0000-0002-1019-8634}}
\authorrunning{Brissman et al.}
\institute{Computer Vision Laboratory, Dept. EE, Link\"oping University, Sweden \and
Saab, Sweden \\
\email{\{emil.brissman, per-erik.forssen, johan.edstedt\}@liu.se}}

\maketitle              

\input{abstract}

\input{introduction}

\input{relatedwork}

\input{method}

\input{experiments}

\input{conclusion}

\input{acknowledgement}

\clearpage
%
%
\bibliographystyle{splncs04}
\bibliography{egbib}
\end{document}

%% file: abstract.tex
\begin{abstract}
A challenge in image based metrology and forensics is intrinsic camera calibration when the used camera is unavailable. The unavailability raises two questions. The first question is how to find the \emph{projection model} that describes the camera, and the second is to detect incorrect models. In this work, we use off-the-shelf extended PnP-methods to find the model from 2D-3D correspondences, and propose a method for model validation. The most common strategy for evaluating a projection model is comparing different models' residual variances—however, this naive strategy cannot distinguish whether the projection model is potentially underfitted or overfitted. To this end, we model the residual errors for each correspondence, individually scale all residuals using a predicted variance and test if the new residuals are drawn from a standard normal distribution. We demonstrate the effectiveness of our proposed validation in experiments on synthetic data, simulating 2D detection and Lidar measurements. Additionally, we provide experiments using data from an actual scene and compare non-camera access and camera access calibrations. Last, we use our method to validate annotations in MegaDepth.
\end{abstract}

%% file: introduction.tex
\begin{figure}[b!]
    \centering
    \includegraphics[width=.308\linewidth, trim={1.4cm 1.5cm 1.0cm 1.0cm}, clip]{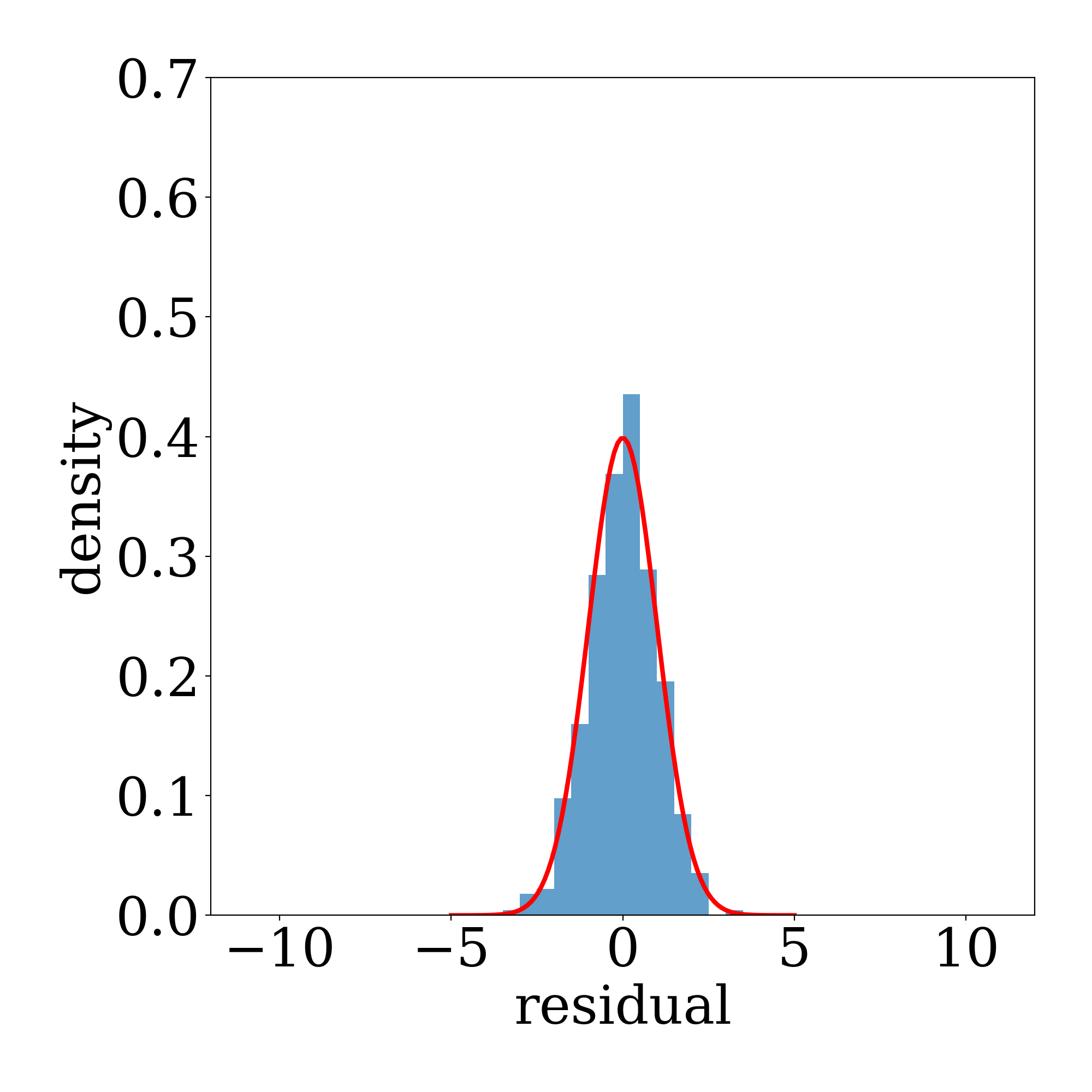}
    \hspace{0.1cm}
    \includegraphics[width=.289\linewidth, trim={2.65cm 1.5cm 1.0cm 1.0cm}, clip]{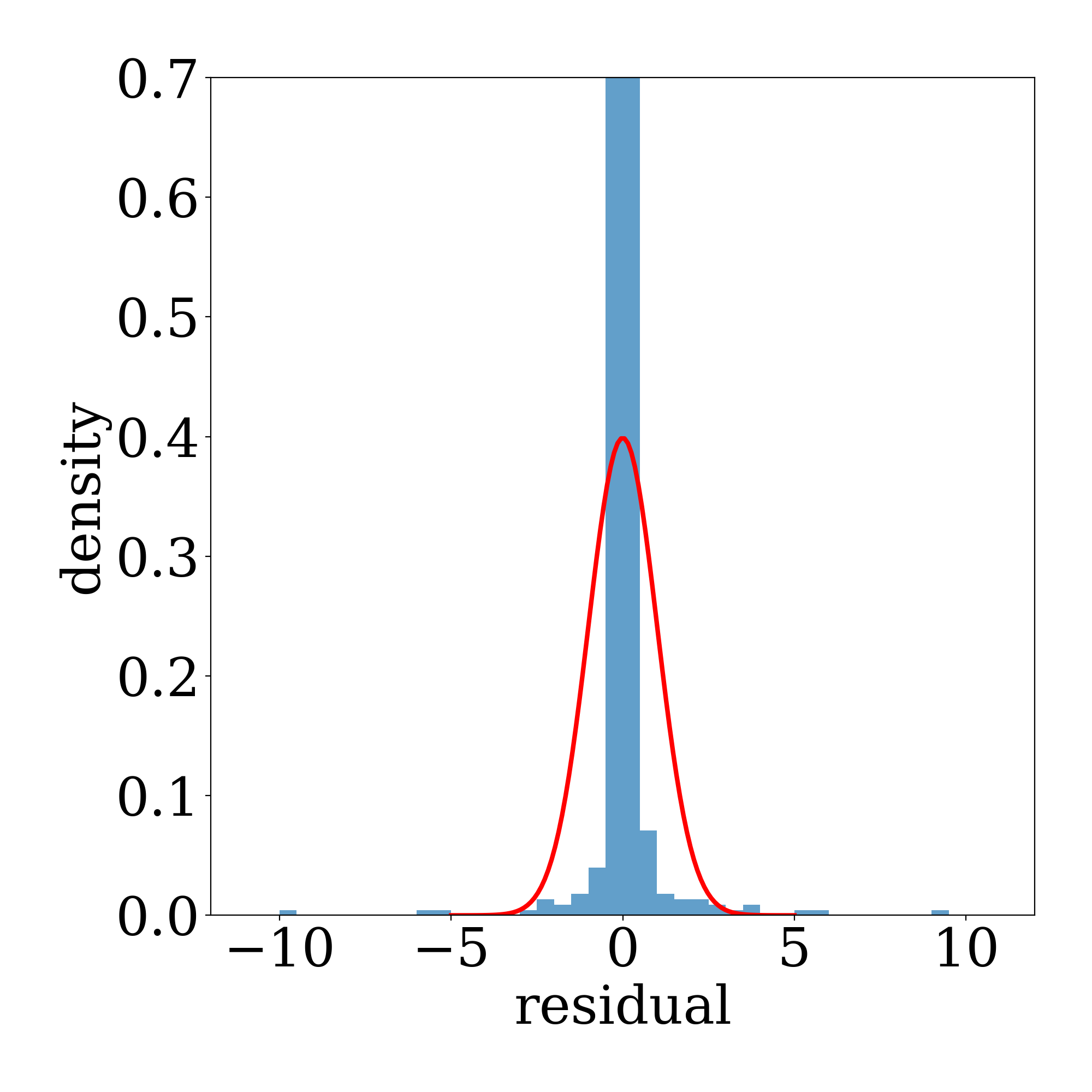}   
    \hspace{0.1cm}
    \includegraphics[width=.289\linewidth, trim={2.65cm 1.5cm 1.0cm 1.0cm}, clip]{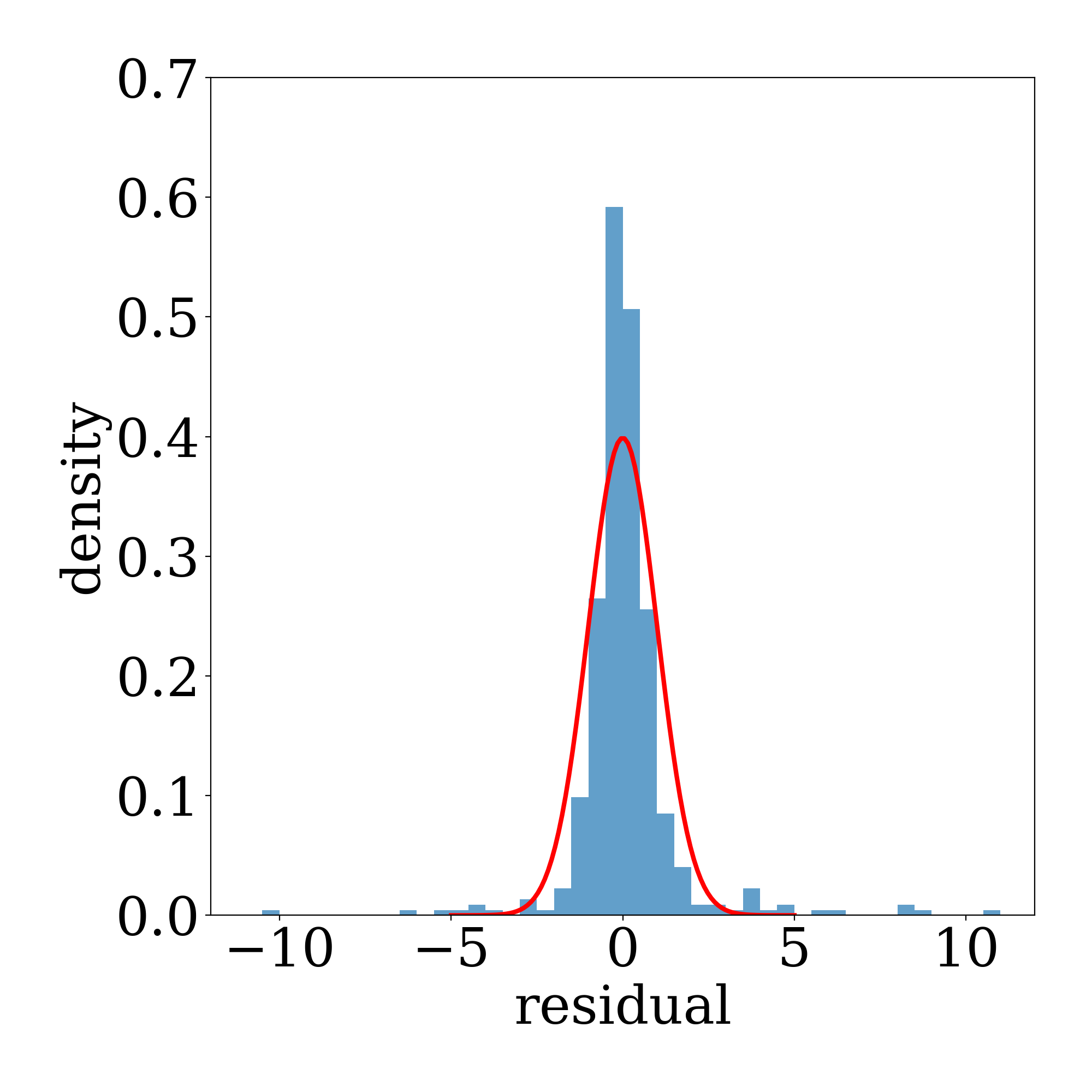}
    \caption{Left: Standardised residuals for a correct model with one distortion parameter using our robust scale estimate. I.e.\ the residuals are not affected by the model error. Middle: Standardised residuals from images under more distortion, for an incorrect model using a non-robust scale estimate. Right: Standardised residuals for an incorrect model, using a robust scale estimate.}
    \label{fig:noise_stuff}
\end{figure}
\section{Introduction}
Intrinsic camera calibration is a fundamental computer vision problem. It involves finding the parameters that allow the conversion of pixel coordinates to bearing angles \cite{hz2003}. It is possible to use the camera for {\it metrology} using a calibration. In the single-view case, metrology means measuring the lengths and angles of objects depicted in an image. As an extension, it is the underpinning of single view 3D reconstruction \cite{immeasure}. Metrology has many applications, including non-contact measurements, sensor fusion, and forensic analysis.

Traditionally, intrinsic calibration is a semi-automatic process, which involves imaging of calibration objects \cite{zhangs,tsai}. Such calibration allows controlled accuracy; however, access to the camera is required. In forensic analysis, the camera is only sometimes available, depending on the received material. Therefore, we aim to facilitate measurements in an image when the camera is unavailable. Using a calibration profile from a camera of the same model often works well, but the accuracy is unknown in this approach and should thus be avoided in forensics.

In the Perspective-n-Point (PnP) problem, the goal is to estimate the camera pose given a set of 2D-3D point correspondences. Early methods assume a calibrated camera, and only estimate translation and rotation parameters~\cite{pnpfishler}. More recent variants of PnP also estimate the intrinsic camera parameters~\cite{absolutepose1,pnp_methods}. These \emph{extended PnP methods} (xPnP) do not require the camera to be available, in contrast to calibration pattern methods \cite{zhangs,tsai}. However, they introduce new challenges such as 2D-3D matching and validation.

In this work, we attend to the validation of camera calibration for forensic metrology applications~\cite{bramble01}. Usually, a model is assumed to be validated if it, on average, has low residuals. However, this approach will not provide any measure of uncertainty in the image plane. Moreover, deriving the uncertainty is challenging because the amount of distortion scales non-linearly with the distance to the camera centre. Thus, we treat noise modelling as a robust regression problem and predict a residual scaling for each 2D-3D correspondence. When the model is correct, we assume the scaled residuals to follow a standard normal distribution (Figure~\ref{fig:noise_stuff}). Next, to verify this assumption, we use a hypothesis test. Simulated data, an indoor scene and MegaDepth~\cite{megadepth}, with annotated cameras depicting different scenes, demonstrate our proposed validation.

\parsection{Contributions} Our contributions are as follows: \textbf{(i)} We propose a method for testing residuals based on variance predictions and standardisation. \textbf{(ii)} We suggest using xPnP methods for unavailable cameras as input to our method, given 2D-3D correspondences. \textbf{(iii)} An empirical estimate of the variance scales residuals poorly. Instead, we propose a predictive noise model to scale individual residuals over the 2D detector and projected 3D noise. \textbf{(iv)} We analyse the effectiveness of our method in quantitative and qualitative experiments and demonstrate its ability to significantly predict incorrect models, also when the mean of the residuals is low.

\subsection{Background Motivation}
At the Swedish National Forensic Center (NFC), the task is to collect information linked to crimes without the possibility of misinterpretations when used in the Swedish court system. At the time of writing, NFC uses the Zhang method~\cite{zhangs} for metrology. However, this commonly accepted practice only applies to images where the camera is available. Lidar scanning is a standard technique in forensic investigations, and in many legal cases, the depicted location is revisited for scanning. Using this working methodology, Olsson~\cite{emilyolsson} first investigated the validation of xPnP methods that forms the basis of this work. In~\cite{emilyolsson}, model correctness is assessed by checking if the empirical mean of the re-projected sample distances is within the two centre quartiles. However, this decision will prefer incorrect models since outliers will expand the decision range.

\subsection{Ethical Consideration}
This work does not concern any police investigations or legal cases. Instead, the method we propose analyses residuals using synthetic data, available benchmark data, and a snapshot of a fictional crime scene provided by a police agency. These are all free from apparent ethical dilemmas.

%% file: relatedwork.tex
\section{Related Work}

\parsection{Semi-automatic Calibration}
Camera calibration is a broad subject found in many areas of industry and research. However, the most common camera calibration practice is to use a printed pattern on a planar surface. This strategy was proposed by Zhang~\cite{zhangs}, who suggested using a checkerboard pattern with equidistant squares of black and white. The inner corners of the pattern form unambiguous features that are easy to find. Detecting several of these features, also called saddle points, between different views allows camera parameters to be estimated. Each detected saddle point in each picture is assigned to its corresponding point on the checkerboard. This set of correspondences represents a series of homographies, determining the intrinsic and extrinsic parameters for one or more cameras. In the case of Tsai~\cite{tsai}, camera calibration depends only on one view of a co-planar checkerboard pattern. The Zhang method~\cite{zhangs}, instead depends on at least three pictures of a planar checkerboard pattern. More recently, deep methods like Li \etal~\cite{cnn_blind} take a single image as input and jointly learn to predict distortion coefficients and optical flow from images with lens-type annotations. However, this problem only concerns visual quality and provides no model accuracy assessment.

\parsection{Perspective-n-Point}
The PnP problem~\cite{pnpfishler} refers to finding a rigid transformation from 2D-3D point matches. That is, to estimate the rotation matrix ${\bf R}$ and the translation vector ${\bf t}$ describing the camera pose in the coordinate system of 
 the 3D points, assuming that the intrinsic parameters never changed during the sampling of a scene. The minimal but ambiguous case, P3P~\cite{gao,pnpsolution,Persson_2018_ECCV} is not considered in this work.

Lepetit \etal~\cite{Lepetit_EPnP} (EPnP), reduced the computational complexity to $O(n)$ operations. Although the convergence is fast, the solution depends on initialization and global convergence is not guaranteed. Later works recognized the need to include intrinsic parameters to generalize application tasks~\cite{pnp_methods}\cite{absolutepose1}. Nakano~\cite{pnp_methods} extends the PnP problem by including intrinsic parameters and dividing the parameter estimation into different stages. Radial distortion and equal focal length horizontally and vertically are assumed, as well as fixating the principal point to the image center. Larsson \etal~\cite{absolutepose1} instead require a minimal correspondence set and add a local optimization step~\cite{lo_ransac}. In our work, we propose a method to validate xPnP methods, for application in forensic analysis, by \emph{Goodness-of-Fit} (GoF) testing between distributions. That is, we do not improve the methods~\cite{pnp_methods} and~\cite{absolutepose1}.

\parsection{Empirical Performance Evaluation}

Works by Wang \etal~\cite{Wang_HypoML} and Thai \etal~\cite{Thai_camident} are related to our work. Wang \etal~\cite{Wang_HypoML} propose a method to test hypotheses about the effect of conceptual changes in deep classification models. That is, if the difference is the probable reason behind the increase in accuracy. Thai \etal~\cite{Thai_camident} propose to identify cameras by raw pixel intensities. Similar to our approach, two quantities parametrize the intensity variance—analog gain, controlled by the camera ISO setting, and electronic noise caused during sensor readout. For an unknown camera the parameters are first estimated and secondly tested against known parameter values (null hypothesis).

\parsection{Goodness of Fit}
The goodness of fit testing is one of the fundamental tasks in statistics. In this work, we focus on normality testing due to normal distributions being a good model for uncertainty in projective geometry~\cite{heuel2004uncertain,forstner2005uncertainty}. Still, our approach could easily be generalized to GoF tests for arbitrary distributions. 

There are a large variety of proposed statistics for normality testing, of which the Kolmogorov-Smirnov (KS)~\cite{Kolmogorov-Smirnov}, D'Agostino-Pearson (DAP)~\cite{dagostino73}, and Shapiro-Wilk (SW)~\cite{shapiro-wilk} tests are well known. These tests all seek to maximize the power,~\ie, minimizing the risk of the null hypothesis being accepted, given that an alternative hypothesis is correct. We discuss those further in Section~\ref{sec:hypothesis-testing} and test all three in Section~\ref{sec:experiments}.

\parsection{Single View Metrology}
Metrology is the study of measurement. In the context of computer vision, single view metrology~\cite{criminisi2000single} involves estimating, \eg, angles and lengths, from a single image. In all metrology, an accurate measure of uncertainty is crucial, and in particular in the forensic setting. Previous work in single view metrology has focused on the undistorted (but often uncalibrated) case~\cite{criminisi2000single}, with model uncertainty assumed to be normally distributed~\cite{criminisi2000single,heuel2004uncertain,forstner2005uncertainty,Brandner10}. At inference time, these uncertainties can be propagated by first order propagation or by Monte Carlo simulation. However, in those works, both the uncertainty estimation and propagation requires \emph{a priori} knowledge of the noise levels and estimation method, and implicitly assume the estimated model is approximately correct.
In contrast to those methods, our approach
\begin{enumerate}
    \item Is estimation agnostic, \ie, we can treat the estimators as black boxes.
    \item Generalises to arbitrary projection models.
    \item Does not implicitly assume that the estimated model is approximately correct.
\end{enumerate}
In particular, perturbation theory, as used in previous work, does not provide a reliable measure of the trustworthiness of the estimated model, it simply provides an approximate measure of the estimation sensitivity to the input. In contrast, our method directly measures trustworthiness by testing the hypothesis of the matches being generated from the estimated model.

%% file: method.tex
\section{Method}
We propose a method that compares observed and expected noise levels. The method takes residual values as input, given a calibration computed from an xPnP method and 2D-3D correspondences. We decompose the residual error for each correspondence as three additive terms: \textbf{(i)} 2D detector noise, \textbf{(ii)} 3D detector noise projected into the image, and \textbf{(iii)} model noise. The expected model noise is zero for the correct model, not affecting the residual distribution in any direction. We describe this in Section~\ref{sec:residual-error-model}. To handle unexpected model noise, Section~\ref{sec:estimation} details robust regression over \textbf{(i)} and \textbf{(ii)} to obtain a scale value for each 2D point. Finally, we assume the scaled residuals are drawn from a standard normal distribution and test this using a GoF test. We motivate our preferred choice of test in Section~\ref{sec:hypothesis-testing}. We consider all points to influence the validation decision and believe this to improve applications in forensic analysis. We begin with an example to get a good intuition of our approach.

\subsection{Motivating Example}
Consider a correct data model $y = x$ and an (incorrect) hypothesis $h_{\text{bad}}: y = x + 0.5x^5$. Under the assumption that $y$ is observed with some Gaussian noise, the residuals $r$ of the true model will be distributed as $\mathcal{N}(0,\sigma_y^2)$. In contrast, the residuals of the incorrect hypothesis are typically \emph{significantly} different from the expected distribution (as shown in Figure~\ref{fig:noise_stuff}). Thus, if $\sigma_y$ is known, a simple hypothesis test is whether $\frac{r}{\sigma_y} \sim\mathcal{N}(0,1)$. However, in real world scenarios $\sigma_y$ is typically not known and needs to be estimated. Since incorrect hypotheses typically contain outliers, it is important with a robust estimate of the noise level. We show these steps in Figure~\ref{fig:noise_stuff}. It is clear that $h_{\text{bad}}$ produces a tailed residual distribution that does not follow the expected Gaussian curve. Hence we can use the KS test~\cite{Kolmogorov-Smirnov} to validate the produced models.

\parsection{Underfittning and Overfitting} It is common for a complex model to be optimized to fit the data $y$ perfectly. We can describe overfitting and underfitting as a constant multiplication of $\sigma_y^2$, yielding residuals distributed as $\mathcal{N}(0, a\sigma_y^2)$. When $a < 1$ the model is overfitted, and when $a > 1$ it is incorrect (underfittning). The following sections describe how we can apply this intuition to validate a camera calibration.

\subsection{Camera Calibration}
\label{sec:camera-calibration}
Calibration fundamentally depends on correspondences of point coordinates. An arbitrary camera, $c$, observes a set of $K$ 3D points $\{{\bf X}_k\}_{k=1}^K$, and a set of corresponding image points $\{{\bf x}_k^c\}_{k=1}^K$. Point sets and correspondences are known $\forall k$, and for each camera. In this work, we consider xPnP based camera calibration using the methods proposed by~\cite{pnp_methods}\ and~\cite{absolutepose1}. Both extrinsic (rotation and translation) and intrinsic (focal length and distortion) parameters in~\eqref{eq:projection} are computed to enable measurement of length and angles in the camera image.

\parsection{Distortion} Depending on the optical system of a camera, small or large displacements of image coordinates can be introduced, called image distortion. Unlike the focal length, which scales the image uniformly, distortion is characterised as scaling the image differently depending on the distance to a distortion centre. The farther the pixels are from the centre of distortion, the more they are distorted. We let the same point represent the distortion and optical centra, which is assumed to be fixed and in the centre of the image.
\begin{equation}
    {\bf y'} = g({\bf y}, \boldsymbol{\theta})
    \label{eq:distortion1}
\end{equation}
We model the distortion as in~\eqref{eq:distortion1}, and let $\boldsymbol{\theta} = \left[ \theta_1, \theta_2, \theta_3\right]$ specify the non-linear distortion terms. When $g$ uses one distortion term, it will be denoted as $\texttt{D(1,0)}$ and as $\texttt{D(3,0)}$ when all three terms are used, according to~\cite{absolutepose1}.

\parsection{Correspondences}
The calibration uses Lidar measurements, which map physical features with high precision by emitting narrow laser beams that are reflected back. Even if the image, whose camera we want to calibrate, and the lidar map are recorded at separate times, there should be enough overlapping features left for calibration. That is, consistent physical properties. Such properties, which are more likely to be consistent, are, for example, those found on buildings, vegetation, paintings, furniture, etc. In practice, correspondences can be of varying quality, making robust estimation a critical importance, when computing an xPnP solution~\cite{absolutepose1}. Therefore, we use only the residuals from correspondences marked as inliers by the model estimator for model validation.

\subsection{Residual error model}
\label{sec:residual-error-model}
Regardless of whether the corresponding coordinates are found by an interest point detector, or whether they are manually annotated by a human, they will suffer from {\it detection noise}. This means that a location estimate $\tilde{\bf x}$ has a residual $\boldsymbol{\epsilon}_{detector}$, compared to the ideal point location $\hat{\bf x}$. This residual is typically modelled as a 2D normal distribution:
\begin{equation}
\boldsymbol{\epsilon}_{detector}=\tilde{\bf x}-\hat{\bf x}\sim\mathcal{N}({\bf 0},\sigma_d^2{\bf I})\,.
\label{eq:residual}
\end{equation}
For a successful calibration, the residual between a detected point, and the projection of the corresponding 3D point using the estimated parameters, should also satisfy \eqref{eq:residual}. In other words:
\begin{equation}
\hat{\bf x}=\textup{proj}_{\boldsymbol{\Theta},{\bf P}}(\hat{\bf X})={\bf K}g(\pi({\bf R}\hat{\bf X}+{\bf t}),\boldsymbol{\theta})\,.
\label{eq:projection}
\end{equation}
Here $\hat{\bf X}$ is the ideal 3D point, and $\pi$ is the pinhole projection. The {\it intrinsic calibration}, $\boldsymbol{\Theta}=({\bf K},\boldsymbol{\theta})$, and the \emph{extrinsic calibration}, ${\bf P}=({\bf R},{\bf t})$ (the camera pose) are of course estimates in practice. We summarize the error caused by the estimation in an additive \emph{modelling noise} term $\boldsymbol{\epsilon}_{model}$. We intend to explain the residuals by detection noise in the image and in the Lidar, and test whether the explanation holds using a test on the residual data,~\eg~by the DAP test \cite{dagostino73}, or by the KS test~\cite{Kolmogorov-Smirnov}, testing the GoF.

For 2D-3D matches, the 3D points are also affected by noise $\boldsymbol{\epsilon}_{3D}$. Thus, \eqref{eq:projection} should be replaced by:
\begin{equation}
\hat{\bf x}-\boldsymbol{\epsilon}_{lidar}-\boldsymbol{\epsilon}_{model}=\textup{proj}_{{\boldsymbol{\Theta}},{\bf P}}(\hat{\bf X}-\boldsymbol{\epsilon}_{3D})\,,
\label{eq:projection2}
\end{equation}
where $\boldsymbol{\epsilon}_{3D}$ is the detection noise in 3D, and $\boldsymbol{\epsilon}_{lidar}$ its projection. By combining \eqref{eq:projection2} with \eqref{eq:residual} we obtain the following residual model:
\begin{equation}
    \boldsymbol{\epsilon}=\boldsymbol{\epsilon}_{detector}+\boldsymbol{\epsilon}_{lidar}+\boldsymbol{\epsilon}_{model}\,.
\label{eq:error_model}
\end{equation}
We model the detection error as in \eqref{eq:residual}, and describe the lidar error model in detail below.

\subsubsection{Lidar error model}

For a Lidar sensor, the 3D noise has both angular and depth components. However, when the camera and 3D-sensor are close to being co-axial, and point in roughly the same direction (i.e.\ ${\bf t}$ is small, and ${\bf R}\approx {\bf I}$ in \eqref{eq:projection}), the depth error becomes irrelevant, and the projection in the image $\boldsymbol{\epsilon}_{lidar}$ is dominated by ${\bf K}$, which is affine. This means that the shape of $\boldsymbol{\epsilon}_{lidar}$ is a simple, but location dependent scaling.

We thus model the projection of the Lidar error $\boldsymbol{\epsilon}_{lidar}$ as:
\begin{equation}
    \boldsymbol{\epsilon}_{lidar} \sim \mathcal{N}(0,\sigma_l^2\textup{diag}(a_x^2,a_y^2))\,,
\label{eq:lidar_error}
\end{equation}
where $\sigma^2_l$ is a noise variance, and $a_x,a_y$ are the noise scalings in horizontal and vertical directions. These depend on the location in the image. To estimate $a_x,a_y$, we can project the current 3D point and its neighbours in pan and tilt directions to obtain:
\begin{align}
    a_{x,k} &=\|\textup{proj}({\bf X}_k)-\textup{proj}({\bf X}_k^P)\|\label{eq:ax}\\
    a_{y,k} &=\|\textup{proj}({\bf X}_k)-\textup{proj}({\bf X}_k^T)\|\label{eq:ay}\,,
\end{align}
where ${\bf X}_k$ is the current 3D point, and ${\bf X}_k^P$, and ${\bf X}_k^T$ are its neighbours in pan and tilt directions.

\subsection{Noise Estimation}
\label{sec:estimation}

The parameters of the detector errors in the model \eqref{eq:error_model} can be fitted to the observed residuals using robust linear regression. However, estimating the variance of $\boldsymbol{\epsilon}_{model}$ is neglected since its observed values are those that will remain in order to test whether the model is incorrect. I.e.\ we assume:
\begin{equation}
    E\left\{\epsilon^2\right\}=\sigma_{d}^2 + \sigma_l^2.
    \label{eq:var_assume}
\end{equation}
By using a common $\sigma_{d}$ for x, and y image residuals, and the aspect ratio model in \eqref{eq:lidar_error} we obtain:
\begin{equation}
    E\left\{\begin{pmatrix}\epsilon_x^2\\\epsilon_y^2\end{pmatrix}\right\}=\begin{pmatrix}
    1 & a_x^2 \\
    1 & a_y^2 \end{pmatrix}\begin{pmatrix} \sigma_{d}^2\\\sigma_l^2\end{pmatrix}.
\end{equation}

We fit these to the observed $K$ residuals, for each camera, $c$, separately, to obtain the regressor parameters $(\sigma_{d},\sigma_{l})$.
\begin{equation}
    \begin{pmatrix}\epsilon_{x,1}^2 & \epsilon_{y,1}^2 & \hdots & \epsilon_{x,K}^2 & \epsilon_{y,K}^2\end{pmatrix}^{\intercal} = \begin{pmatrix}
    1 & 1 & \hdots & 1 & 1\\
    a_{x,1}^2 & a_{y,1}^2 & \hdots & a_{x,K}^2 & a_{y,K}^2
    \end{pmatrix}^{\intercal}\begin{pmatrix} \sigma_{d}^2\\\sigma_l^2\end{pmatrix}
    \label{eq:normal_eq}
\end{equation}
In practice, we do not use linear regression by solving the normal equations to~\eqref{eq:normal_eq}, but use robust regression using IRLS~\cite{irls} with initial weights $1/p(\boldsymbol{\epsilon}_k | \sigma=5.0)$. We can now obtain standardised residuals:
\begin{equation}
\tilde{\boldsymbol{\epsilon}}_k=\begin{pmatrix}
\epsilon_{x,k}/\sigma_{k,x}\\
\epsilon_{y,k}/\sigma_{k,y}\end{pmatrix} = \begin{pmatrix}
\epsilon_{x,k}/\sqrt{\sigma_{d}^2+a_{x,k}^2\sigma^2_l}\\
\epsilon_{y,k}/\sqrt{\sigma_{d}^2+a_{y,k}^2\sigma^2_l}\end{pmatrix}.
\label{eq:stdres}
\end{equation}

\subsection{Hypothesis Testing}
\label{sec:hypothesis-testing}
When the modelling error is low, the standardised residuals in~\eqref{eq:stdres} should pass a statistical test, such as, \eg, the KS test. We can thus use the test to check whether the calibration worked for a particular set of 2D-3D correspondences. More formally, we test the {\bf null hypothesis} $\mathcal{H}_0$ : {\it The standardised residuals~\eqref{eq:stdres} are distributed explicitly according to a standard normal}, against $\mathcal{H}_1$: {\it at least one value does not match that distribution}. Related to this classical approach is that the data we are testing is random, so the test decision is random too, which means there is still a small probability of an incorrect decision. Nevertheless, tests are useful to detect low model errors and thus further validate the calibration.

\parsection{Evidence} The approach involves comparing the samples (residuals) with a statistical model under $\mathcal{H}_0$, where a test statistic measures the discrepancy between the data and the model. To this end, we use the KS test~\cite{Kolmogorov-Smirnov} and compute a \emph{p-value}, measuring the error size of rejecting $\mathcal{H}_0$. Commonly, when the \emph{p-value} is below $5\%$, $\mathcal{H}_0$ can be rejected in favour of $\mathcal{H}_1$. That is, the error probability is sufficiently low. However, this probability does not directly infer confidence for the data distributed as a standard normal.

Other tests also calculate a \emph{p-value} to test the normality of data. For example, the DAP test~\cite{dagostino73} sums the discrepancies from a skewness test and a kurtosis test into a single \emph{p-value}. Skewness is the asymmetry about the mean, and kurtosis is the measure of the "tailedness". Although parametric tests are preferable to non-parametric ones, and the SW test is one of the more powerful~\cite{Ghasemi2012}, we believe their null hypotheses to be non-directional, where a broader chance of normality is possible, leading to unstable decisions.

%% file: experiments.tex
\section{Experiments}
\label{sec:experiments}
We first evaluate the proposed method for testing a calibration using 2D-3D correspondences on synthetic data simulating detection and Lidar errors. Next, we provide results on a real scenario using Lidar measurements and compare this with a semi-automatic calibration. Last, we analyse a large-scale dataset using our method.

\begin{figure}[t!]
    \centering
    \includegraphics[width=0.48\linewidth, trim={0.5cm 17.8cm 13.0cm 0.7cm}, clip]{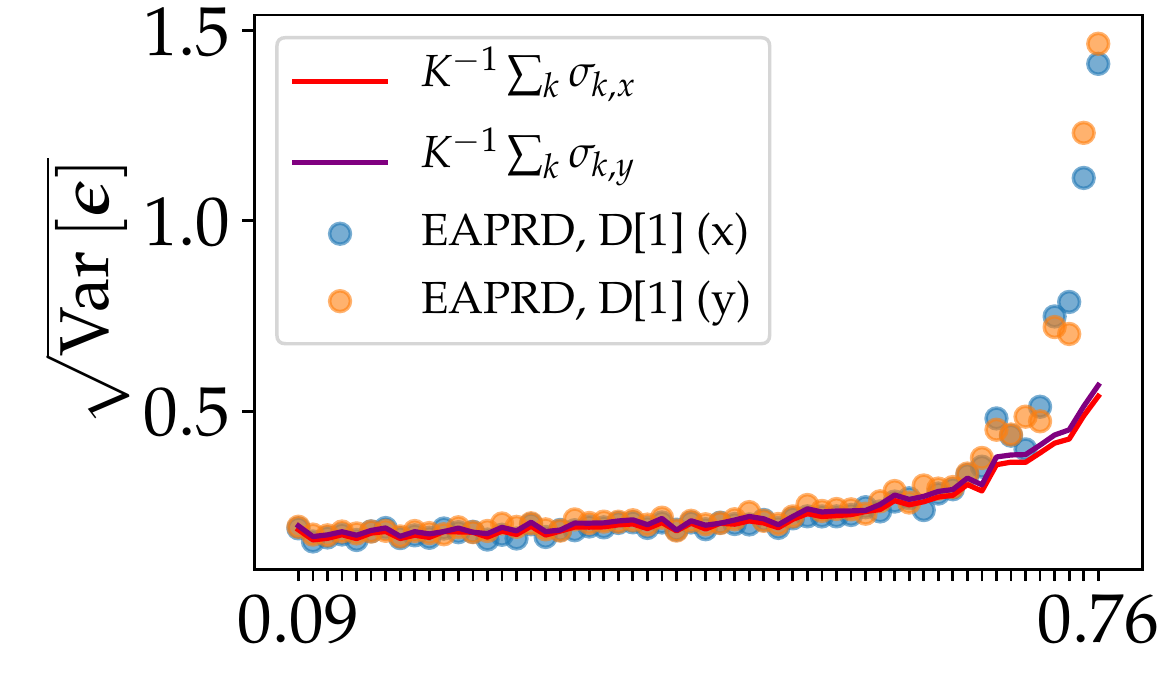}
    \includegraphics[width=0.445\linewidth, trim={1.75cm 17.8cm 12.5cm 0.7cm}, clip]{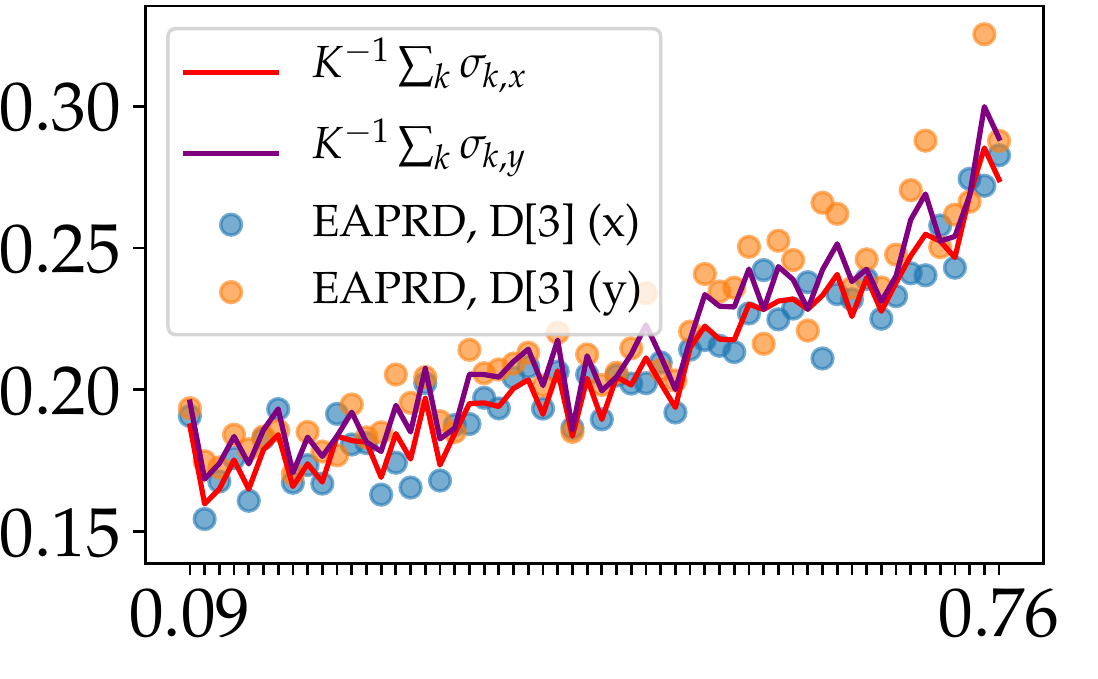}\\
    \includegraphics[width=0.48\linewidth, trim={13.0cm 17.8cm 0.5cm 0.7cm}, clip]{figures/sim/simulated_all_scale01_EAPRD_D1_kstest.pdf}
    \includegraphics[width=0.435\linewidth, trim={14.0cm 17.8cm 0.5cm 0.7cm}, clip]{figures/sim/simulated_all_scale01_EAPRD_D3_kstest.pdf}\\
    \includegraphics[width=0.5\linewidth, trim={12.5cm 17.8cm 0.5cm 0.7cm}, clip]{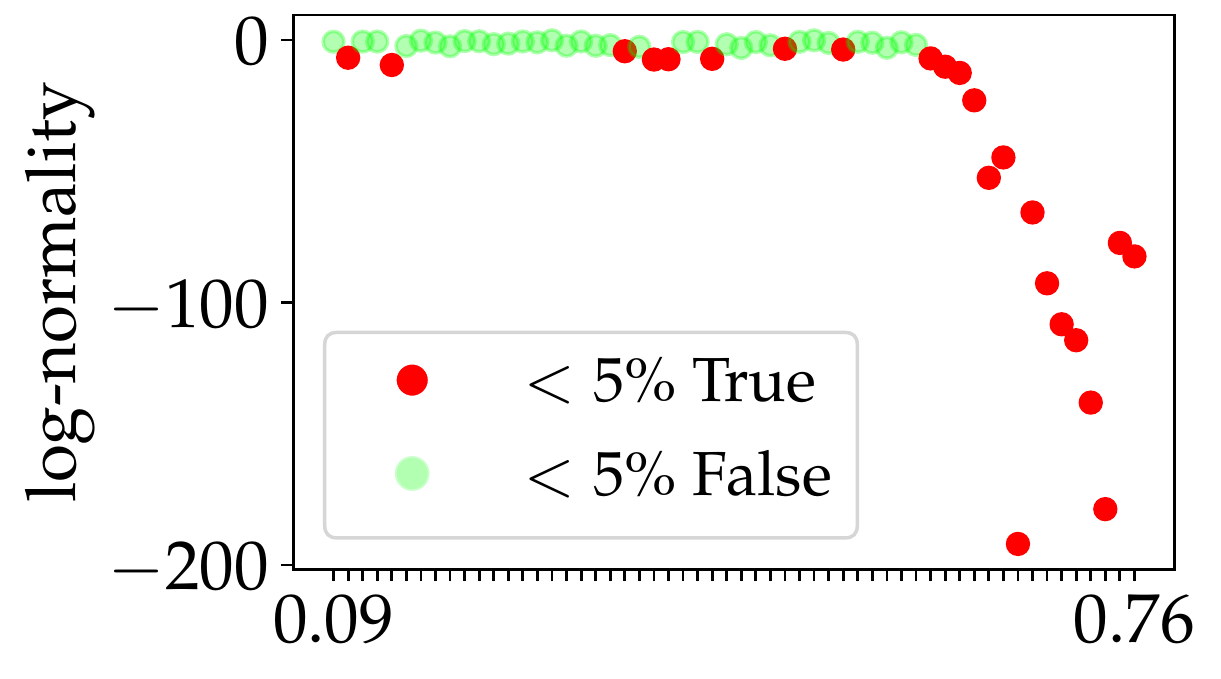}
    \includegraphics[width=0.455\linewidth, trim={14.0cm 17.8cm 0.0cm 0.7cm}, clip]{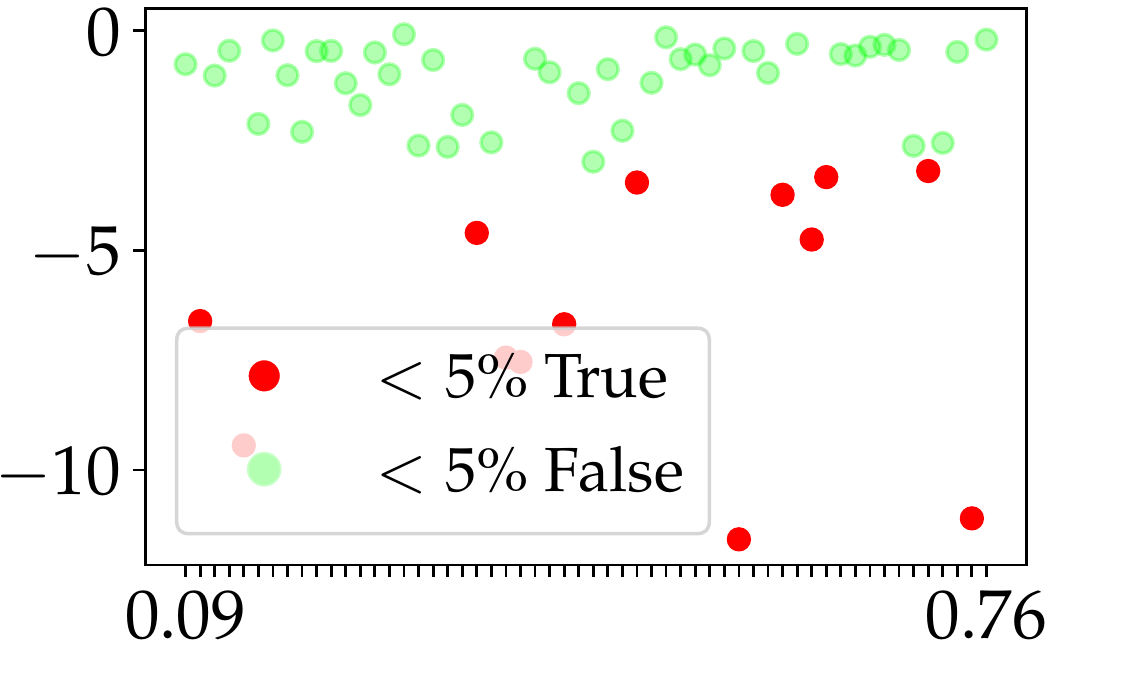}\\
    \includegraphics[width=0.5\linewidth, trim={12.5cm 16.8cm 0.5cm 0.7cm}, clip]{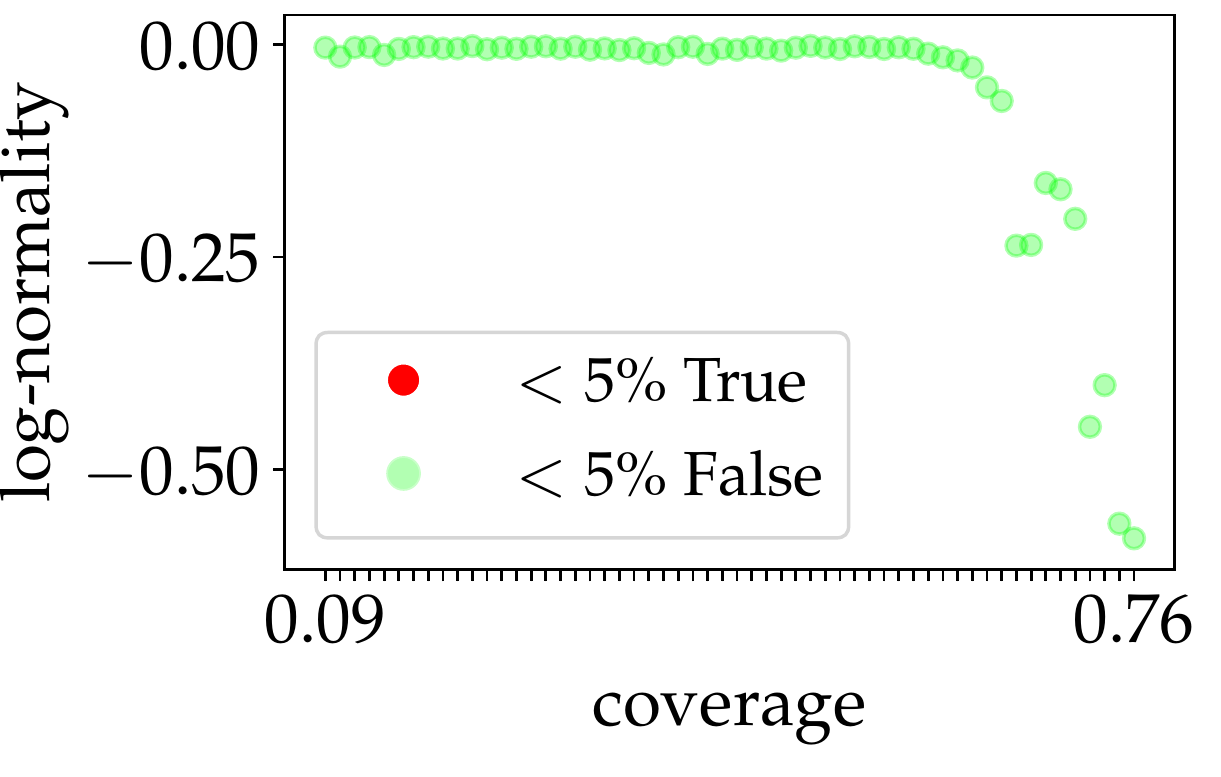}
    \includegraphics[width=0.47\linewidth, trim={14.0cm 16.8cm 0.0cm 0.7cm}, clip]{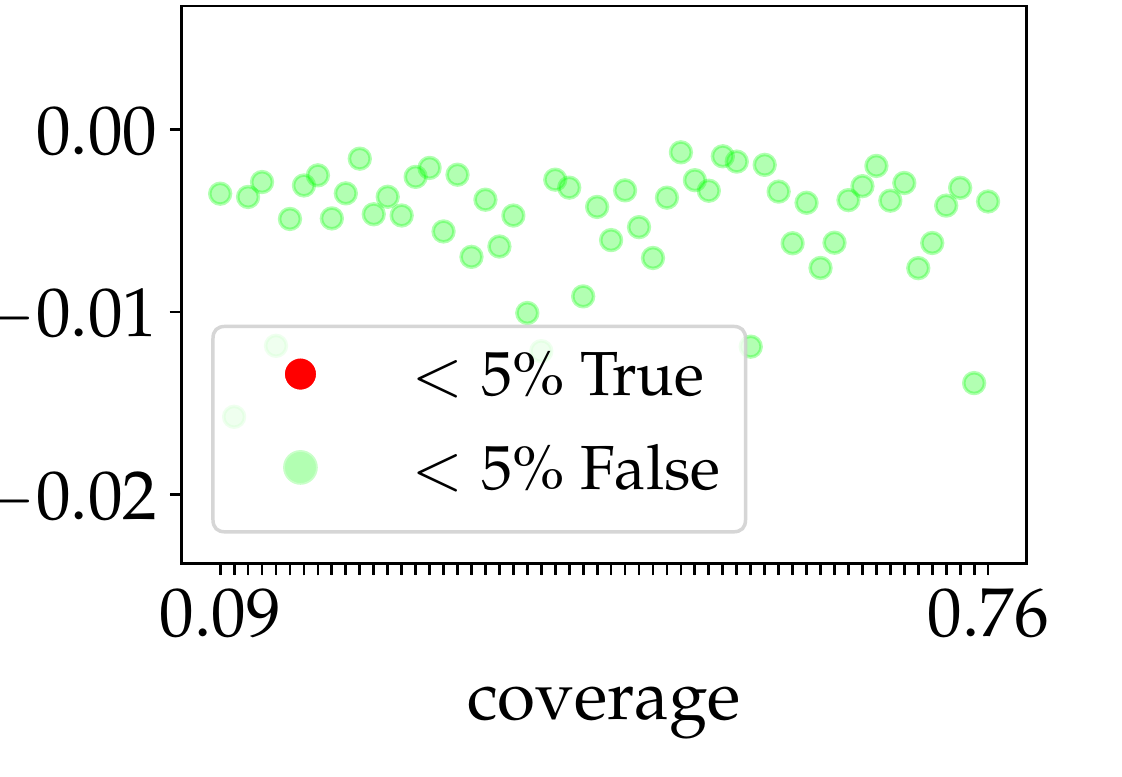}\\
    \caption{Using residuals computed from $56$ sets of simulated 2D-2D correspondences, we show the results for the incorrect ${\tt D(1,0)}$ model (left) and the correct ${\tt D(3,0)}$ model (right). We sort the correspondence sets in ascending order, using the area of the 2D points' convex hull (\emph{coverage}). The second, third and fourth rows show the outcome of the KS~\cite{Kolmogorov-Smirnov}, DAP~\cite{dagostino73} and SW~\cite{shapiro-wilk} tests at level $5\%$. When the model is correct, the standard deviation is low (first row), and our predicted variance follows the corresponding empirical value. The KS test rejects images under an incorrect model, while accepting images under the correct model. In contrast,~\cite{dagostino73} is too strict, while~\cite{shapiro-wilk} is too permissive.}
   \vspace{2mm}
   \label{fig:sim_d1}
\end{figure}

\begin{table}[b!]
    \centering
    \resizebox{0.7\columnwidth}{!}{%
    \begin{tabular}[b]{c c c c c c c c c c c c}\hline
    Model & $f_x$ & $f_y$  & $c_x$ & $c_y$ & $k_1$ & $k_2$ & $k_3$ & $\texttt{std(\textrm{\emph{axis223m}})}$ & $\texttt{std(\textrm{\emph{axisp3364}})}$ \\\hline
    $\textsc{m}_1$ & \checkmark &            &            &            & \checkmark & & & 2.99 & 2.91\\
    $\textsc{m}_2$ & \checkmark &            & \checkmark & \checkmark & \checkmark & & & 2.96 & 2.73 \\
    $\textsc{m}_3$ & \checkmark &            &            &            & \checkmark & \checkmark & \checkmark & 3.03 & 2.95\\
    $\textsc{m}_4$ & \checkmark &            & \checkmark & \checkmark & \checkmark & \checkmark & \checkmark & 3.19 & 2.52\\
    $\textsc{m}_5$ & \checkmark & \checkmark & \checkmark & \checkmark & \checkmark & \checkmark & \checkmark & 0.81 & 0.19\\
    \hline
    \end{tabular}}
    \caption{Projection models used with the Zhang method~\cite{zhangs}. Models $\textsc{m}_1$ and $\textsc{m}_3$ use the same number of parameters as the PnP methods $\texttt{D(1,0)}$~\cite{absolutepose1}, $\texttt{D(3,0)}$~\cite{absolutepose1} and $\texttt{D(3,0)}$~\cite{pnp_methods}. The standard deviation on a set of test images, determines how accurately the two cameras have been calibrated, \emph{axis223m} and \emph{axisp3364}. $\textsc{m}_5$ is the most accurate model. Models $\textsc{m}_1$-$\textsc{m}_4$ (approximately) share the same standard deviation, although models $\textsc{m}_4$ and $\textsc{m}_5$ only differ with one parameter. Our method instead decides $\textsc{m}_1$ as incorrect but $\textsc{m}_3$ as a plausible model, still usable for metrology.}
    \label{tab:cam_models}
\end{table}

\subsection{Synthetic data}
\label{sec:synthetic}
We implemented the simulator in OpenCV~\cite{opencv_library} and will provide code upon publication. We aim to render a fictitious checkerboard pattern with equidistant squares of black and white into a camera $c$ with a small angular rotation maintaining the image centre as its viewpoint at a distance ${\boldsymbol t}$. The pattern contains equally many saddle points (inner checkerboard corners) vertically as horizontally ($15 \times 15$). The simulator iterates three main tasks to render each image, which is presented next. \textbf{(i)} The projection model has fixed intrinsic parameters according to $\texttt{D(3,0)}$~\cite{absolutepose1}. That is, $f=800$ and $\boldsymbol{\theta} = \left[-0.0684,  0.0100, 0.0006 \right]$. We let the distortion centre coincide with the image centre. Rotation parameters are randomly sampled in the range $\pm 15^{\circ}$ relative to the z-axis. The translation can also be random, but in our generated synthetic dataset, we move the pattern closer and closer to the camera. \textbf{(ii)} Next, we smooth the image (to avoid aliasing), add image noise and interpolate it to size $1600\times1600$. A saddle point detector~\cite{opencv_library} locates the 2D position of these features with sub-pixel precision. We observed the position error to lie within a small range of $0.03$ pixels. This corresponds to $\sigma_d$ in \eqref{eq:var_assume}.\textbf{(iii)} To simulate the Lidar, we add noise on the corresponding 3D points in all images. We transform the noise such that it lives on the sphere with origin ${\boldsymbol t}^c$ and radius $||{\boldsymbol t}^c||$. On the sphere, the noise magnitude is dependent on ${\boldsymbol t}^c$, and in the vertical direction, the noise is always $90\%$ lower compared to the horizontal direction. This reweighing aims to simulate the resolution in the Lidar array, and it replaces \eqref{eq:ax} and~\eqref{eq:ay}, which are used on real datasets. This gives us the resolution aspect, which is used as \emph{explanatory variables} in the estimate,~\eqref{eq:normal_eq}.

\begin{figure}
    \centering
    \includegraphics[width=\linewidth, trim={5cm 5cm 5cm 5cm}, clip]{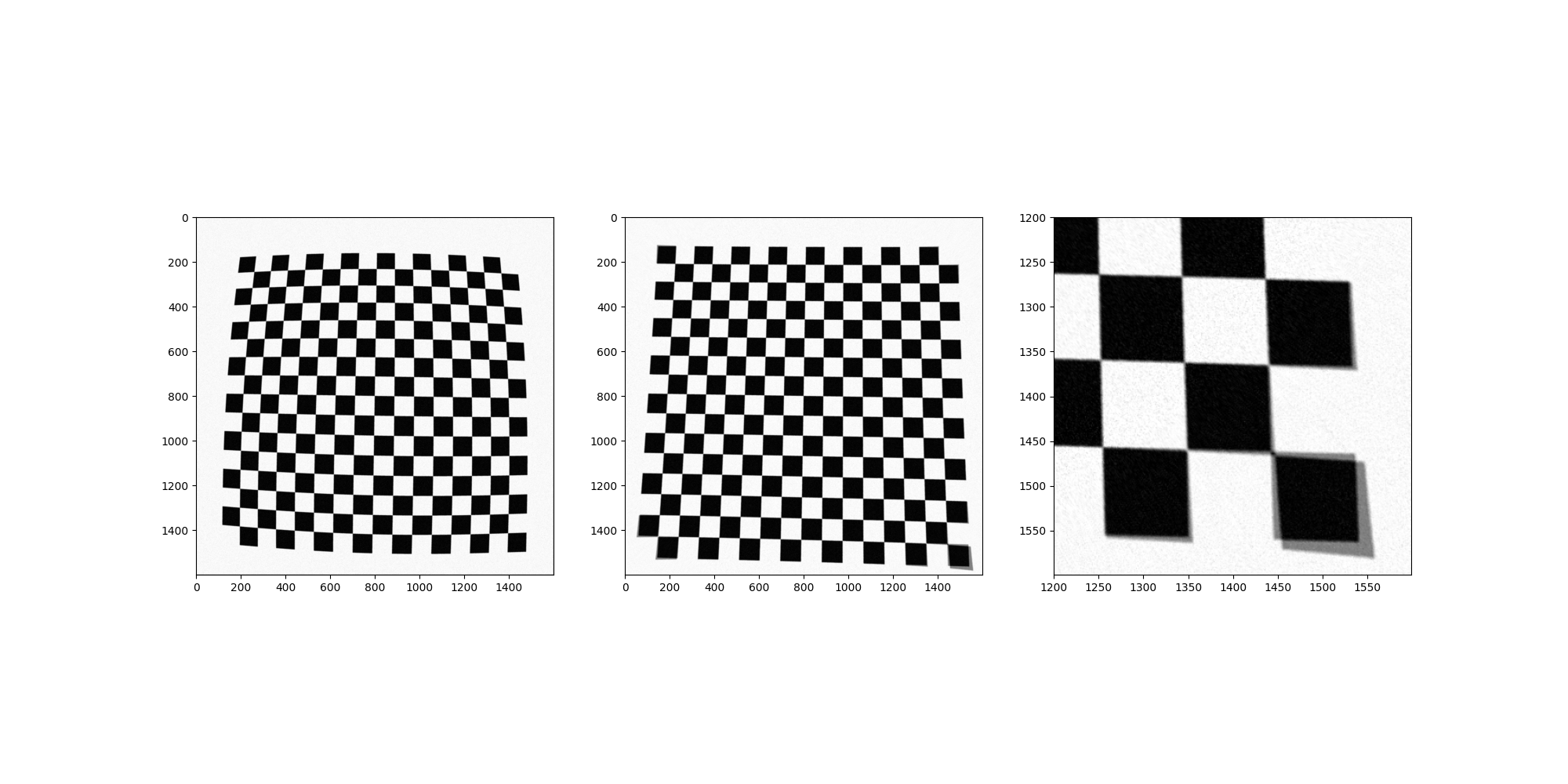}
    \caption{Incorrect models can have small residual errors, but \emph{Godness-of-Fit} testing exposes them. Left: A distorted image from which the model is estimated. Middle: Undistorted image using true model (black) and wrong model (grey) overlaid. Even though the estimated model is incorrect, when detector errors are small the residual errors are often small, hence simply checking the standard deviation of the residuals is insufficient. Right: Outliers indicate model failure. Given known outlier-free correspondences, deviations from the expected noise distribution expose incorrect models.}
    \label{fig:undistortion-errors}
\end{figure}

\parsection{Results}
In Figure~\ref{fig:sim_d1}, we show the output of our method on models $\texttt{D(1,0)}$~\cite{absolutepose1} and $\texttt{D(3,0)}$~\cite{absolutepose1} for a set of synthetic 2D-3D correspondences, generated under the $\texttt{D(3,0)}$ model. These are sorted according to the increasing spatial spread of correspondences. The closer the points are to the image edge, the more they are affected by the distortion (Figure~\ref{fig:undistortion-errors}). The first row in Figure~\ref{fig:sim_d1} shows the empirical and estimated standard deviations over all points per image, and rows 2-4 whether the scaled residuals are sufficient evidence to reject the null hypothesis for three different tests. Given in the second row of Figure~\ref{fig:sim_d1} is the output of the KS test for both $\texttt{D(1,0)}$~\cite{absolutepose1} and $\texttt{D(3,0)}$~\cite{absolutepose1}. To the left, in the same row, our method rejects the images with correspondences more uniformly spread over the entire image at level $5\%$ using $\texttt{D(1,0)}$~\cite{absolutepose1}. Making the same test using a projection model with more parameters fits the data more accurately, as indicated by both the low standard deviation and the test. We also test the scaled residuals in the third and fourth rows using~\cite{dagostino73}, and~\cite{shapiro-wilk}, respectively. While these tests are parametric, they test for any normal distribution. Compared to the proposed KS test, this leads to false positives and negatives, see rows 3-4 of Figure~\ref{fig:sim_d1}. In the second column, the stronger $\texttt{D(3,0)}$ projection model is tested. In most images, $\mathcal{H}_0$ can not be rejected as expected due to the simulated data conforming to $\texttt{D(3,0)}$~\cite{absolutepose1}. However, the output does not reveal the tests' differences in this case. Thus,~\cite{Kolmogorov-Smirnov} generally leads to more accurate decisions.

\begin{table}[t!]
  \centering
  \resizebox{0.6\columnwidth}{!}{%
    \begin{tabular}[b]{l c c c c c}\hline
    PnP Method & DT & $\sqrt{\textrm{Var}\left[\boldsymbol{\epsilon}\right]}$ & KS\cite{Kolmogorov-Smirnov} & DAP\cite{dagostino73} & SW\cite{shapiro-wilk} \\\hline
    EAPRD~\cite{absolutepose1} & $\texttt{D(1,0)}$ & 2.23 & \checkmark & \checkmark & \\
    EAPRD~\cite{absolutepose1} & $\texttt{D(3,0)}$ & 0.98 &            & \checkmark & \\
    PNPRF~\cite{pnp_methods}   & $\texttt{D(3,0)}$ & 1.05 &            & \checkmark & \\
    \hline
    \end{tabular}}
  \caption{The results for the \emph{axis223m} camera. For each PnP method, with distortion type DT, we report the empirical standard deviation and whether $\mathcal{H}_0$ can be rejected at level $5\%$ using the KS, DAP or SW  test.}
  \label{tab:test_axis223m}
\end{table}
\begin{table}[t!]
  \centering
  \resizebox{0.6\columnwidth}{!}{%
    \begin{tabular}[b]{l c c c c c}\hline
    PnP Method & DT & $\sqrt{\textrm{Var}\left[\boldsymbol{\epsilon}\right]}$ & KS\cite{Kolmogorov-Smirnov} & DAP\cite{dagostino73} & SW\cite{shapiro-wilk} \\\hline
    EAPRD~\cite{absolutepose1} & $\texttt{D(1,0)}$ & 1.71 & \checkmark & \checkmark & \\
    EAPRD~\cite{absolutepose1} & $\texttt{D(3,0)}$ & 1.47 &            &            & \\
    PNPRF~\cite{pnp_methods}   & $\texttt{D(3,0)}$ & 1.08 &            &            & \\
    \hline
    \end{tabular}}
  \caption{The results for the \emph{axisp3364} camera. For each PnP method, with distortion type DT, we report the empirical standard deviation and whether $\mathcal{H}_0$ can be rejected at level $5\%$ using the KS, DAP or SW test.}
  \label{tab:test_axisp3364}
\end{table}

\subsection{Lidar Measurements}
\label{sec:exp_lidar}
\begin{figure*}[t!]
    \centering
    \includegraphics[width=0.24\linewidth]{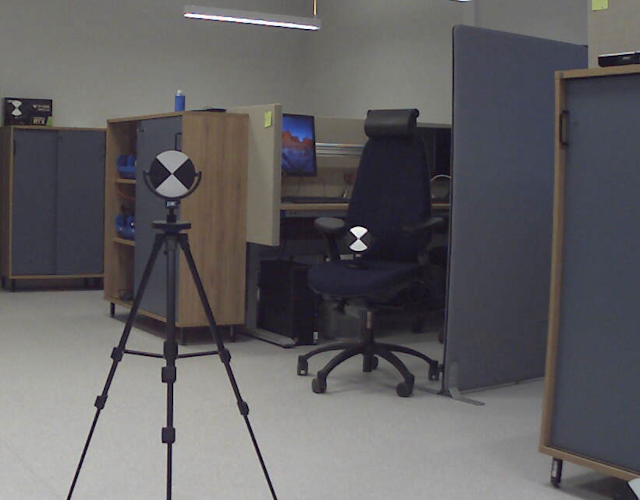} \includegraphics[width=0.24\linewidth]{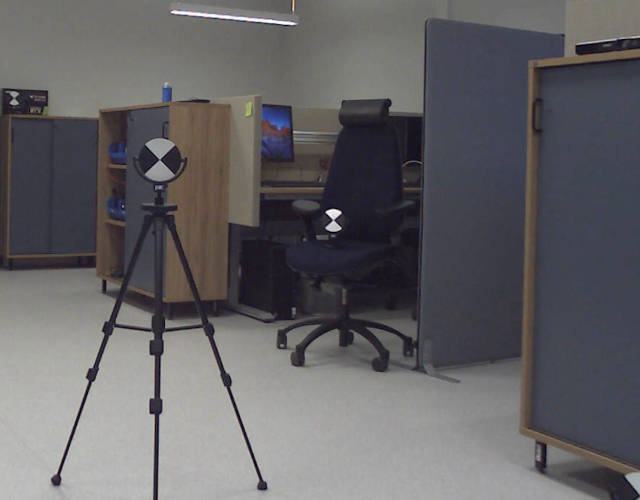}
    \includegraphics[width=0.24\linewidth]{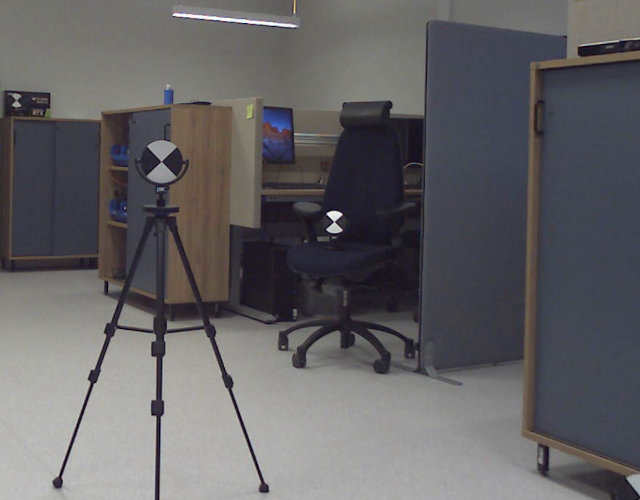}
    \includegraphics[width=0.24\linewidth]{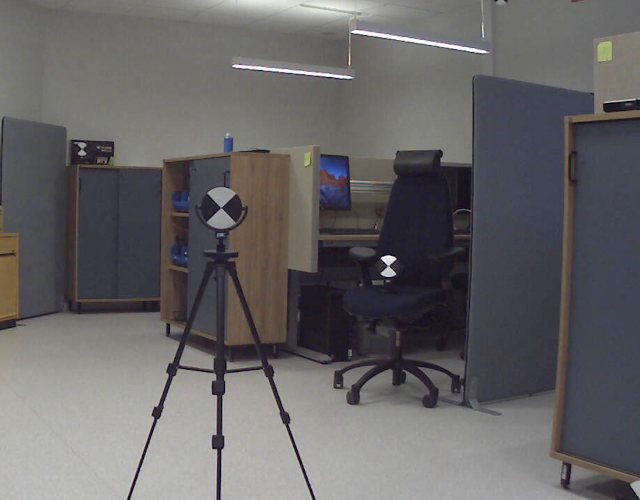}\vspace{0.8mm}\\
    \includegraphics[width=0.24\linewidth]{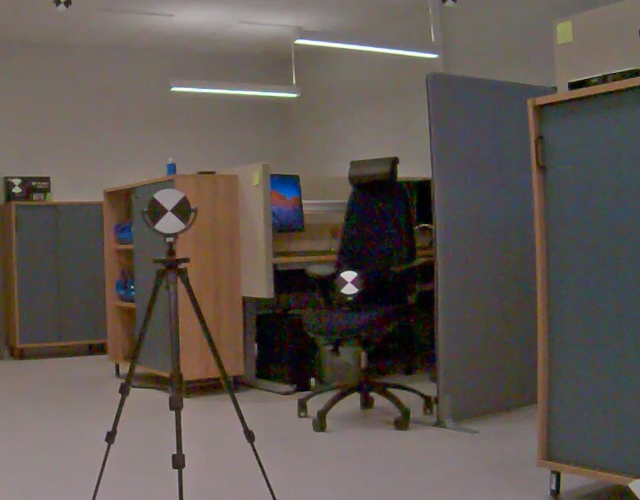} \includegraphics[width=0.24\linewidth]{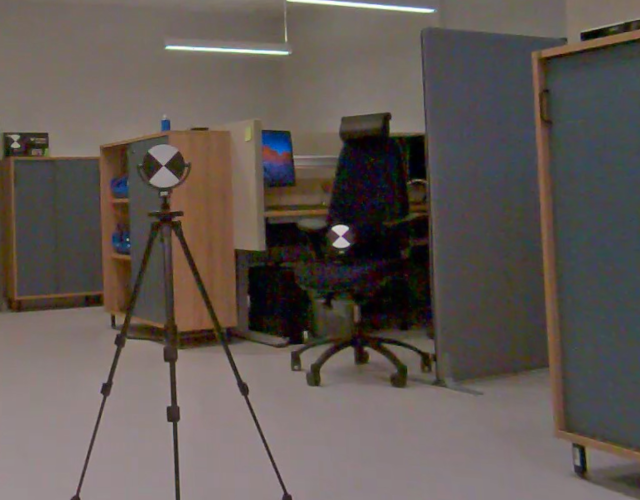}
    \includegraphics[width=0.24\linewidth]{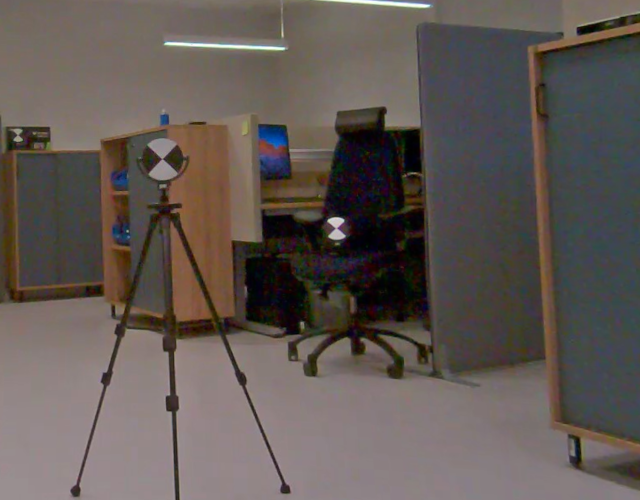}
    \includegraphics[width=0.24\linewidth]{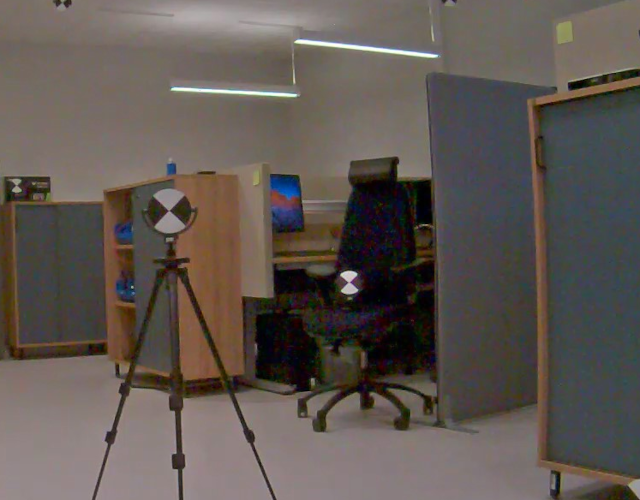} 
    \caption{The first column shows two images of two cameras, \emph{axis223m} and \emph{axisp3364}, respectively. In the second and third columns, the undistortion looks to be visually removed for both $\texttt{D(1,0)}$~\cite{absolutepose1}, and $\texttt{D(3,0)}$~\cite{absolutepose1}. Our method correctly detects $\texttt{D(1,0)}$~\cite{absolutepose1} as incorrect for both cameras (Tables~\ref{tab:test_axis223m} and~\ref{tab:test_axisp3364}). The undistorted images in the fourth column are visually similar to their original, but this is not detected. For more details, see Section~\ref{sec:exp_lidar}.} 
   \label{fig:real_cam_scenario}
\end{figure*}

Next, we compare our method using images from two real cameras, \emph{axis223m} and \emph{axisp3364}, respectively, and a 3D point cloud from a \emph{Leica RTC360} scanner, with semi-automatic calibration. The second camera offers lower-quality images than the first, which is visible in Figure~\ref{fig:real_cam_scenario}. There is also no verified annotation for the cameras; in practice, there is none, and the camera can be inaccessible. The cameras instead depict a scene such that their optical axes have a relatively small angle to the 3D point cloud coordinate system's z-axis, similar to the simulations in Section~\ref{sec:synthetic}.

\parsection{Semi-automatic} For semi-automatic calibration, we collect images for both cameras using a checkerboard pattern. The pattern has $6 \times 6$ saddle points. We split the set of images and estimate the model parameters on the first set using the Zhang method~\cite{zhangs} implemented in~\cite{opencv_library}. Then, we calibrate using different projection models where $\textsc{m}_5$, in Table~\ref{tab:cam_models}, achieves the lowest standard deviation on the second set of images (test). A factor of almost 4 differs between the accuracy of models $\textsc{m}_4$ and $\textsc{m}_5$. In Table~\ref{tab:cam_models}, the number of model parameters differs by one.

\parsection{Without Camera Access} We show in Tables~\ref{tab:test_axis223m} and~\ref{tab:test_axisp3364} that $\texttt{D(1,0)}$~\cite{absolutepose1} is incorrect compared to $\texttt{D(3,0)}$~\cite{absolutepose1} and $\texttt{D(3,0)}$~\cite{pnp_methods} using the KS test on 24 manually annotated correspondences. Each of the annotated 3D points is visible in both cameras and projected to consistent features,~\eg~corners. Similar to simulation, we observe that the parametric tests contradict each other and are thus infeasible for our application. While models $\textsc{m}_1$ to $\textsc{m}_4$, in Table~\ref{tab:cam_models}, obtain higher standard deviations using~\cite{zhangs}, models $\textsc{m}_1$ and $\textsc{m}_3$ are equvivalent to the models used from~\cite{absolutepose1} and~\cite{pnp_methods}, and thus there is possibility that $\textsc{m}_5$ is overfitted.

Finally, we found that the computed distortion parameters of $\texttt{D(3,0)}$~\cite{pnp_methods} were all zero, shown in the rightmost column of Figure~\ref{fig:real_cam_scenario}. To our knowledge,~\cite{pnp_methods} divides the xPnP problem into subproblems. In the subproblem that solves distortion, we can't find a condition on $\boldsymbol{\theta}$ preventing the \emph{normal equations} from giving the trivial solution. Thus, our method can not make the correct decision to either reject $\mathcal{H}_0$ or not based on residuals from $\texttt{D(3,0)}$~\cite{pnp_methods}.

\subsection{Structure-from-Motion}
\begin{figure*}
    \centering
    \includegraphics[height=3.3cm, trim={1.3cm 1.3cm 1.3cm 1cm}, clip]{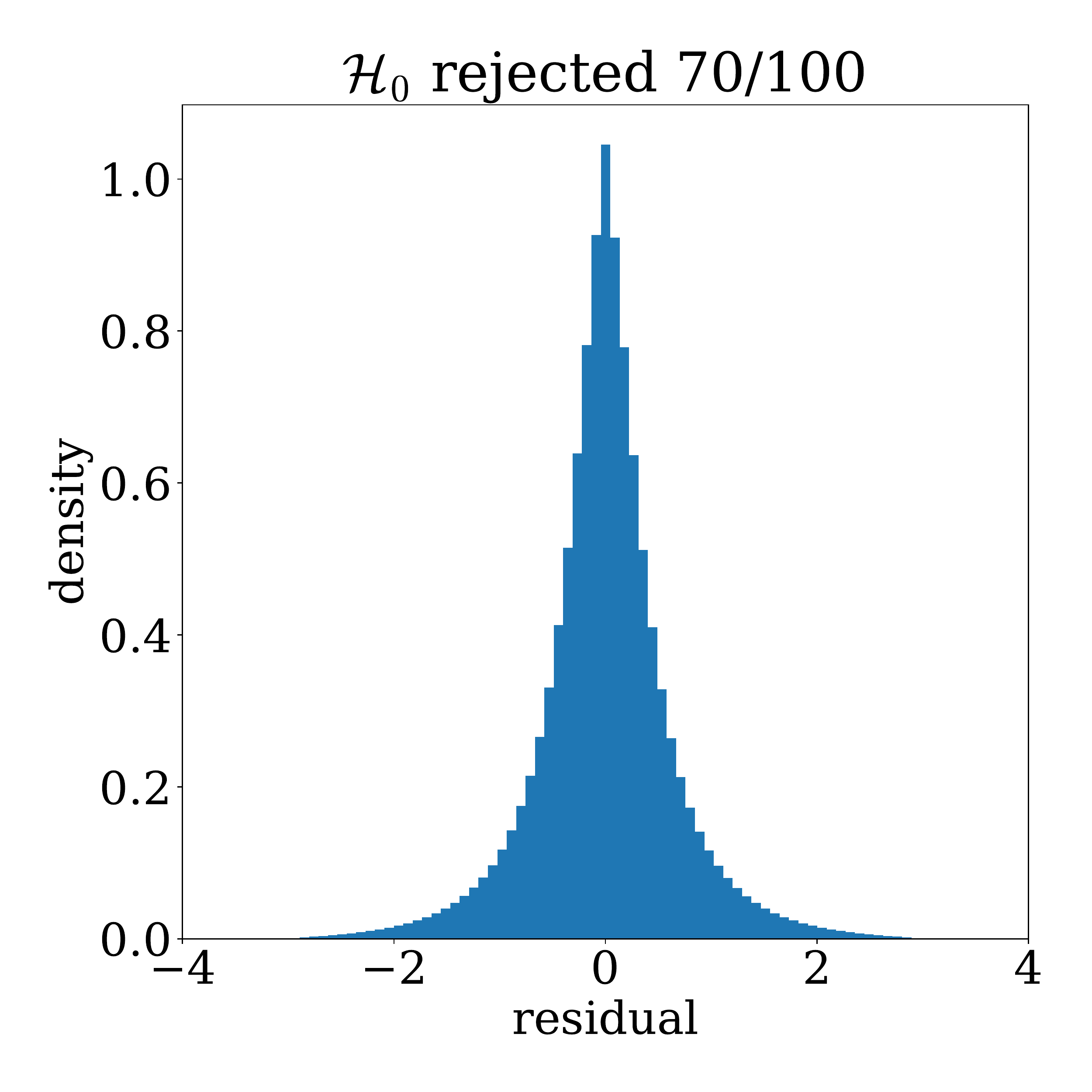} \includegraphics[height=3.3cm]{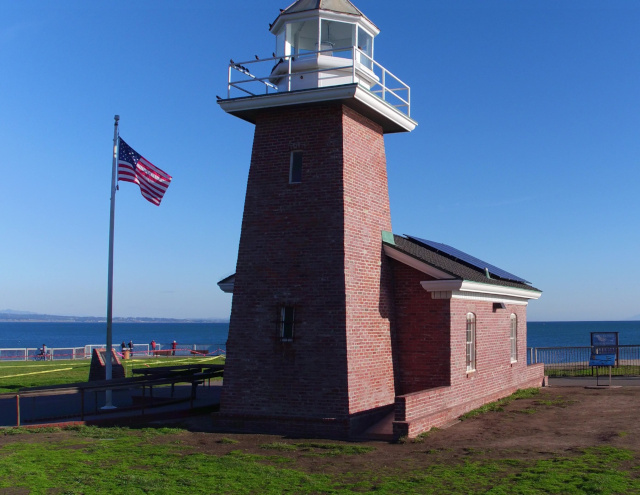}
    \includegraphics[height=3.3cm]{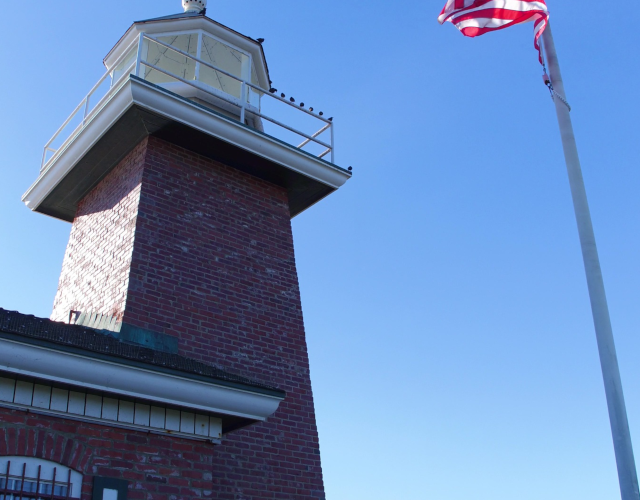}
    \caption{Left: Density plot of residuals from 1000 images in all scenes, on which the annotation in MegaDepth depends. It is unlikely residuals will be high for images in~\cite{megadepth} measuring a good performance. However, our proposed method tests each image and rejects the null hypothesis, $\mathcal{H}_0$, on 70 out of 100 images. Middle: Example of when $\mathcal{H}_0$ can not be rejected, and the \emph{simple radial} projection model is suitable. Right: Example of when our method rejects $\mathcal{H}_0$. As can be seen,~\eg~on the flagpole to the right, the images are distorted.} 
   \label{fig:sfm}
\end{figure*}

In this experiment, we use our proposed method on annotations computed from a Structure-from-Motion (SfM) pipeline to get a broader insight into its effectiveness. To this end, we use 1000 images from each scene of MegaDepth~\cite{megadepth}. This dataset contains many scenes with 2D-3D correspondences and camera intrinsic and extrinsic parameters given. The SfM pipeline, COLMAP~\cite{schoenberger2016sfm}\cite{schoenberger2016mvs}, estimates the annotation parameters of the widely used benchmark for state-of-the-art comparison. In the dataset, the assumed projection model, a \emph{simple radial}, models a single focal length, one distortion parameter, the distortion centre, rotation and translation. The histogram to the left in Figure~\ref{fig:sfm} shows that residuals are overall low. However, in 70 out of 100 images, our method rejects the null hypothesis at level $5\%$. The two images in the middle and to the right, in Figure~\ref{fig:sfm}, show when $\mathcal{H}_0$ can not be rejected at level $5\%$ and when $\mathcal{H}_0$ is rejected in favour of $\mathcal{H}_1$. We can thus assume mostly overfitted projection models in~\cite{megadepth}.

%% file: conclusion.tex
\section{Conclusion}
We suggested that metrology applications in forensic analysis use xPnP methods and use our proposed method to validate the calibration without camera access. The method formulation processes a single image, estimating a robust scaling of each correspondence and tests if the scaled set of residuals is drawn from a standard normal distribution. We demonstrate via qualitative and quantitative experiments that the KS test is most suitable and provide further insight from an extensive collection of annotated cameras.

Although we are sufficiently confident that the test can determine models as incorrect with a small margin of error, the challenge remains to infer confidence in the image measurements. A test is not a classification, and the \emph{p-value} does not imply measurement confidence. However, when rejection of the null hypothesis is not possible at the acceptable error level, our error model explicitly provides the expected measurement errors over the image. Depending on the number of correspondences, we can get local estimates of expected measurement error from our assumptions of normally distributed residuals. Therefore, our method is a useful tool for xPnP camera calibration.

%% file: acknowledgement.tex
\subsection*{Acknowledgement} \emph{This work was partially supported by the Wallenberg AI, Autonomous Systems, and Software Program (WASP) funded by the Knut and Alice Wallenberg Foundation; and the computations were enabled by the Berzelius resource provided by the Knut and Alice Wallenberg Foundation at the National Supercomputer Centre; and a point cloud of a realistic scene was provided by the Swedish National Forensic Centre (NFC).}